\documentclass[runningheads,a4paper]{llncs}

\usepackage{amsmath}
\usepackage{graphicx}            
\usepackage[caption=false]{subfig}
\usepackage{amssymb}             
\usepackage{amsfonts}            
\usepackage{enumitem}            
\usepackage{multirow}            
\usepackage{siunitx}             
\usepackage{url}                 
\usepackage{xspace}              
\usepackage[T1]{fontenc}         
\usepackage{hyperref}            
\usepackage{wrapfig}
\usepackage{todonotes}
\usepackage{units}
\usepackage{subfig}
\usepackage{lipsum}
\usepackage[bottom]{footmisc}    
\usepackage{booktabs,dcolumn}    
\usepackage[firstpage=true]{background}

\newcolumntype{d}[1]{D{.}{.}{#1}} 

\usepackage[%
square,        
comma,         
numbers,       
sort&compress, 
]{natbib}

\graphicspath{{images/}}
\DeclareGraphicsExtensions{.pdf,.png,.jpg,.jpeg}

\newcommand{\seclabel}[1]{\label{sec:#1}}

\newcommand{\figlabel}[1]{\label{fig:#1}}

\newcommand{\figref}[1]{Fig.~\ref{fig:#1}\xspace}

\newcommand{\cmnew}{CM740\xspace}

\usepackage[firstpage=true]{background}
\newcommand\copyrighttext{%
	\parbox{\textwidth}{
		\footnotesize
		In: RoboCup 2019, Robot World Cup XXIII. LNCS 11531, pp. 631-645, Springer, 2019.
	}
}

\SetBgContents{\copyrighttext}
\SetBgScale{1}
\SetBgColor{black}
\SetBgAngle{0}
\SetBgOpacity{1}
\SetBgPosition{current page.north}
\SetBgVshift{-2.5cm}
\SetBgHshift{3mm}

\begin{document}

\mainmatter

\title{RoboCup 2019 AdultSize Winner NimbRo: \\
	Deep Learning Perception, In-Walk Kick, Push Recovery, and Team Play Capabilities
}
\titlerunning{RoboCup 2019 AdultSize Winner NimbRo}

\author{Diego Rodriguez, Hafez Farazi, Grzegorz Ficht, Dmytro Pavlichenko, Andr\'{e} Brandenburger, Mojtaba Hosseini, Oleg Kosenko, Michael Schreiber, Marcel Missura, and Sven Behnke}
\authorrunning{Rodriguez, Farazi, Ficht et al.}

\institute{Autonomous Intelligent Systems, Computer Science, Univ.\ of Bonn, Germany\\
	\url{http://ais.uni-bonn.de}
	\url{rodriguez@ais.uni-bonn.de},
}

\maketitle
\begin{abstract}
	Individual and team capabilities are challenged every year by rule changes and the increasing performance of the soccer teams at RoboCup Humanoid League.
	For RoboCup 2019 in the AdultSize class, 
	the number of players (2 vs. 2 games) and the field dimensions were increased,
	which demanded for team coordination and robust visual perception and localization modules.
	In this paper, we present the latest developments that lead team NimbRo to win the soccer tournament, 
	drop-in games, technical challenges and the Best Humanoid Award of the RoboCup Humanoid League 2019 in Sydney.
	These developments include a deep learning vision system, in-walk kicks, step-based push-recovery, and team play strategies.
\end{abstract}

\section{Introduction}

\begin{figure}[!b]
\centering
\includegraphics[height=49mm]{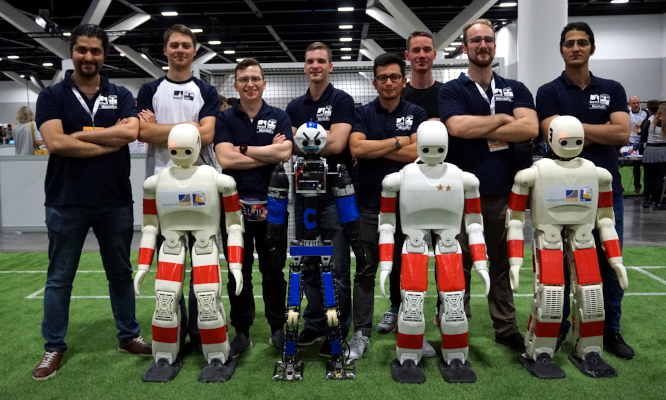}\hspace{1.3mm}%
\caption{Humanoid AdultSize team NimbRo at RoboCup 2019 in Sydney.}
\figlabel{nimbro_team}
\vspace{-2ex}
\end{figure}

The Humanoid League contributes to the goal of beating the human soccer world champion by 2050 by gradually making the game rules more FIFA alike.
Additionally, individual and team skills are also encouraged by a set of technical challenges.
This year, for the AdultSize class, the teams were allowed to be composed by two team players. 
Correspondingly, the field dimensions were updated to 14$\times$\unit[9]{m}.
These modifications pose several challenges in terms of perception (further away balls and goalposts to be detected), locomotion (longer distances demanding for a faster gait), localization (robust line detection and state estimation), and team play (coordination between players).
This paper presents our recent developments to address these modifications and shows their performance in the competition.
These developments include in-walk kicks, a step-based push recovery approach, a vision system based on deep learning and team play strategies.

Our robots won all AdultSize 2019 competitions, namely the soccer tournament, the drop-in games and the technical challenges.
Additionally, the NimbRo team was given with Best Humanoid award of the Humanoid League.
In RoboCup 2019, we used our fully open-source 3D printed humanoid platform NimbRo-OP2(X)~\cite{ficht2017nop2,ficht2018nimbro},
shown in \figref{nimbro_team} with our human team members.
We released a video of the 2019 competition highlights \footnote{RoboCup 2019 NimbRo highlights video: \url{https://youtu.be/ITe-seb4PsA}}.

\section{Robot Platforms}
\seclabel{robot_platforms}
During the competition three different robots have been used --- NimbRo-OP2 (\figref{nimbro_team} first from right), NimbRo-OP2X (\figref{nimbro_team} first from left) and Copedo (\figref{nimbro_team} second from left).
These platforms lead team NimbRo to win all possible competitions last year in the AdultiSize League of RoboCup 2018~\cite{Farazi2019NimbRo}.
Despite the visible differences in the kinematic structure and outer appearance,
there is a fundamental level of similarity between the platforms. 
The joints of all robots are actuated with Robotis Dynamixel actuators. 
These are controlled through a Robotis \cmnew microcontroller board, which also incorporates 
an IMU with a 3-axis gyroscope and accelerometer. 
For visual perception, a Logitech C905 USB camera in combination with a wide-angle lens was used. 
\subsubsection{NimbRo-OP2(X).}
\seclabel{nop2x}
NimbRo-OP2~\cite{ficht2017nop2} and NimbRo-OP2X~\cite{ficht2018nimbro} are our self-develo\-ped humanoid robots, where both the hardware\footnote{NimbRo-OP2X hardware: \url{https://github.com/NimbRo/nimbro-op2}} and software\footnote{NimbRo-OP2X software: \url{https://github.com/AIS-Bonn/humanoid_op_ros}} components are completely open-source.
Although the platforms share a similar design and name, there is a number of differences between them. 
With the same height of \SI{1.34}{m}, the robots place on the lower end of the AdultSize class requirements.
Their 3D printed plastic structure mainly contributes to the low weights of \SI{17.5}{kg}~(OP2) and \SI{19}{kg}~(OP2X). 
Both robots share a similar joint layout with 18 Degrees of Freedom~(DoF), with 5~DoF in a parallel kinematics arrangement per leg, 3~DoF per arm, and 2 DoF actuating the head. 
A Shuttle X1 Gaming computer with an Nvidia GTX 1060 and an Intel Core i7-7700HQ CPU was mounted inside the hollow trunk of the NimbRo-OP2.
For the NimbRo-OP2X we increased the available space in the trunk to fit a standard Mini-ITX motherboard,
with a i7-8700T CPU and a GTX 1050 Ti GPU.
Other significant differences
With the NimbRo-OP2X, a new type of Dynamixel X actuator---the XH540---was used. 
Equipped with thicker gears and a fully metal casing, the servomotors are more durable and reliable, compared to the MX-106 from the NimbRo-OP2. Both robots use external gearing to produce
the required torque in the leg roll and yaw joints. Initially, they were custom-milled out of brass for the OP2. Due to weight and procurement factors, the OP2X utilizes 
3D-printed double-helical gears, which are fast to manufacture.
\subsubsection{Copedo.}
\seclabel{copedo}
Copedo was built using milled carbon composite and aluminum parts. 
These provide the necessary rigidity, while keeping the total weight down. 
Initially, Copedo was built (in 2012) as a TeenSize robot. 
With the introduction of one vs. one games in the AdultSize class in 2017, we have rebuilt him to have a weight 
of \SI{10.1}{kg} and a height of \SI{131}{cm}~\cite{ficht2018Grown}. 
Dynamixel EX-106+ were chosen to power the 5~DoF legs. 
The legs are additionally equipped with tension springs, which allow for energy storage during locomotion. 
The 1~DoF arms and 2~DoF neck use RX-64 servos, due to lower torque and speed requirements for their joints. 
A small, light-weight and efficient Intel NUC NUC7I7BNH (i7-7567U CPU) computer was fitted into Copedo to complete the build. 

\section{Deep Learning Visual Perception}
\seclabel{perception}
Our visual perception pipeline improved significantly since RoboCup 2018.
Thanks to our new unified perception convolutional neural network (NimbRoNet2), we now can reliably perceive the environment in extremely low and very bright lighting condition.
The visual perception system can recognize soccer-related objects, including a soccer ball, 
field boundaries, robots, line segments, and goalposts through the usage of texture, shape, brightness, and color information.

Our deep-learning-based visual perception system is robust against brightness, viewing angles, and lens distortions.
To achieve this, we designed a unified deep convolutional neural network to perform object detection and pixel-wise classification with one forward pass.
After post-processing, we managed to outperform our previous non-deep learning approach to soccer vision~\cite{farazi2015} as well as our previous deep-learning-based model \cite{ficht2018nimbro}.
Our perception system is also able to track~\cite{farazi2016real} and identify our robots~\cite{Farazi2017b}.

The system has two output heads; one for object detection, and the other for pixel-wise segmentation. 
The detection head gives the location of the ball, robots, and goalposts. 
The segmentation head is for line and field detection. 
Our model uses an encoder-decoder architecture similar to pixel-wise segmentation models like SegNet~\cite{badrinarayanan2015segnet}, and U-Net~\cite{ronneberger2015u}. Due to computational limitations and the necessity of real-time perception, we have made several adaptations,
e.g., using a shorter decoder than the encoder. 
Thus, the number of parameters has been reduced for the cost of losing fine-grained spatial information which can be alleviated using sub-pixel post-processing. 
To minimize annotation efforts, we utilized transfer-learning.
A pre-trained ResNet-18
is chosen as the encoder.
Since ResNet was originally designed for recognition tasks, we removed the Global Average Pooling (GAP) and the fully connected layers in the model.
Transpose-convolutional layers are used for up-sampling the representations. 
To use location-dependent features, we used newly proposed location-dependent convolutional layer~\cite{AziziFarazi2018}.
In order to limit the number of parameters used, a shared learnable bias between both output heads is implemented. 
The proposed visual perception architecture
is illustrated in \figref{net}.

\begin{figure}[h]
	\vspace{-3ex}
	\centering
	\includegraphics[width=0.65\linewidth,angle=90,origin=c]{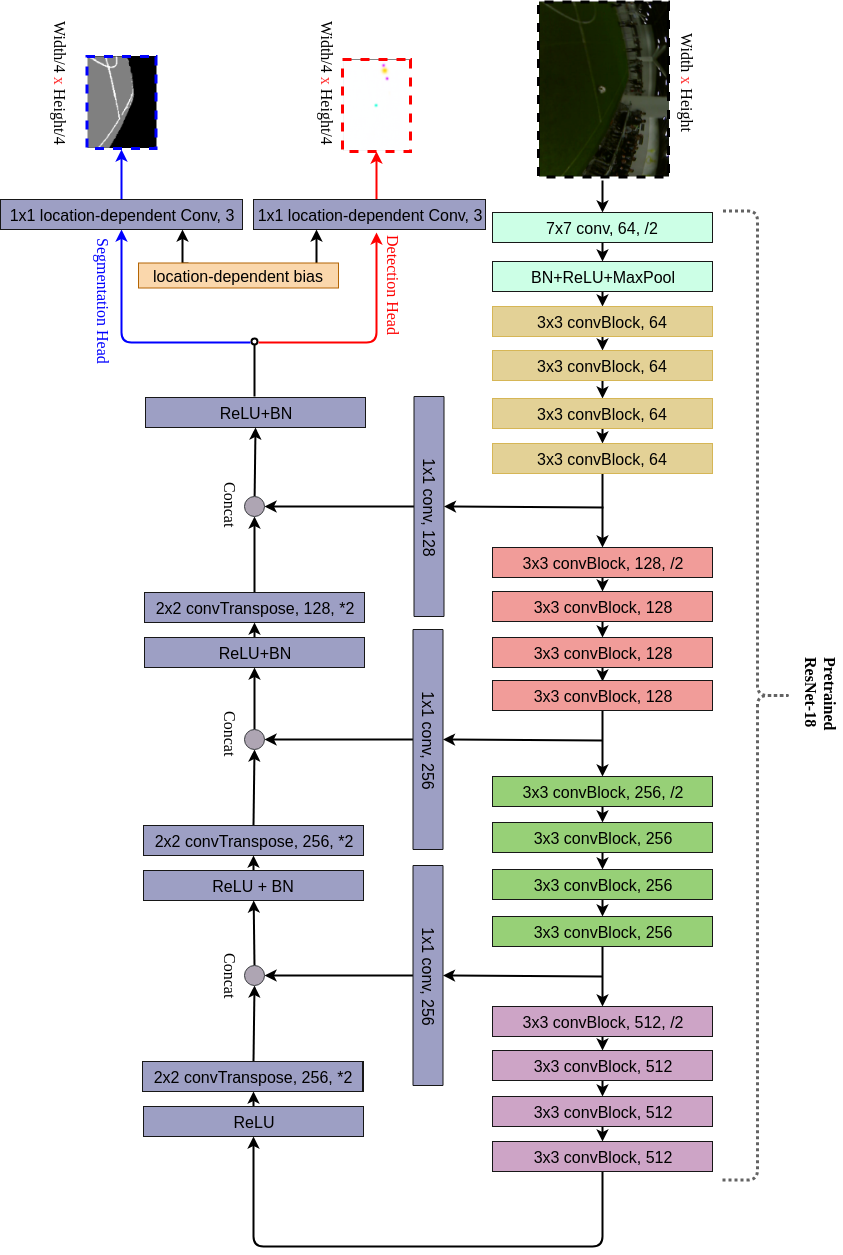}
	\vspace{-14ex}
	\caption{NimbRoNet2 architecture. Similar to ResNet, each convBlock consists of two convolutional layers followed by batch-norm and ReLU activations.
		For simplicity, residual connections in ResNet are not depicted. Note that instead of a convolutional layer we used a location-dependent convolution in the last layer.}
	\figlabel{net}
	\vspace{-3ex}
\end{figure}

Different losses were used for different network heads. For detection head, similar to SweatyNet~\cite{schnekenburger2017detection}, 
the mean squared error is employed.
The target is constructed by Gaussian blobs around the ball center and bottom-middle points of the goalposts and robots. 
In contrast to last year model, NimbRoNet2 uses a bigger radius for robots with the intuition that annotating a canonical center point is more difficult,
thus a bigger radius would less penalize the network for not outputting the exact human labels.
In the classification head, we used pixel-wise Negative Log Likelihood.
We also added Total Variation loss to the output of all result channels except the line segmentation channel.
Total Variation loss encouraged blob response thus helped to have less false positives, especially in field detection. 

\begin{figure}[h]
	\vspace{-2ex}
	\centering
	\includegraphics[width=1.0\linewidth]{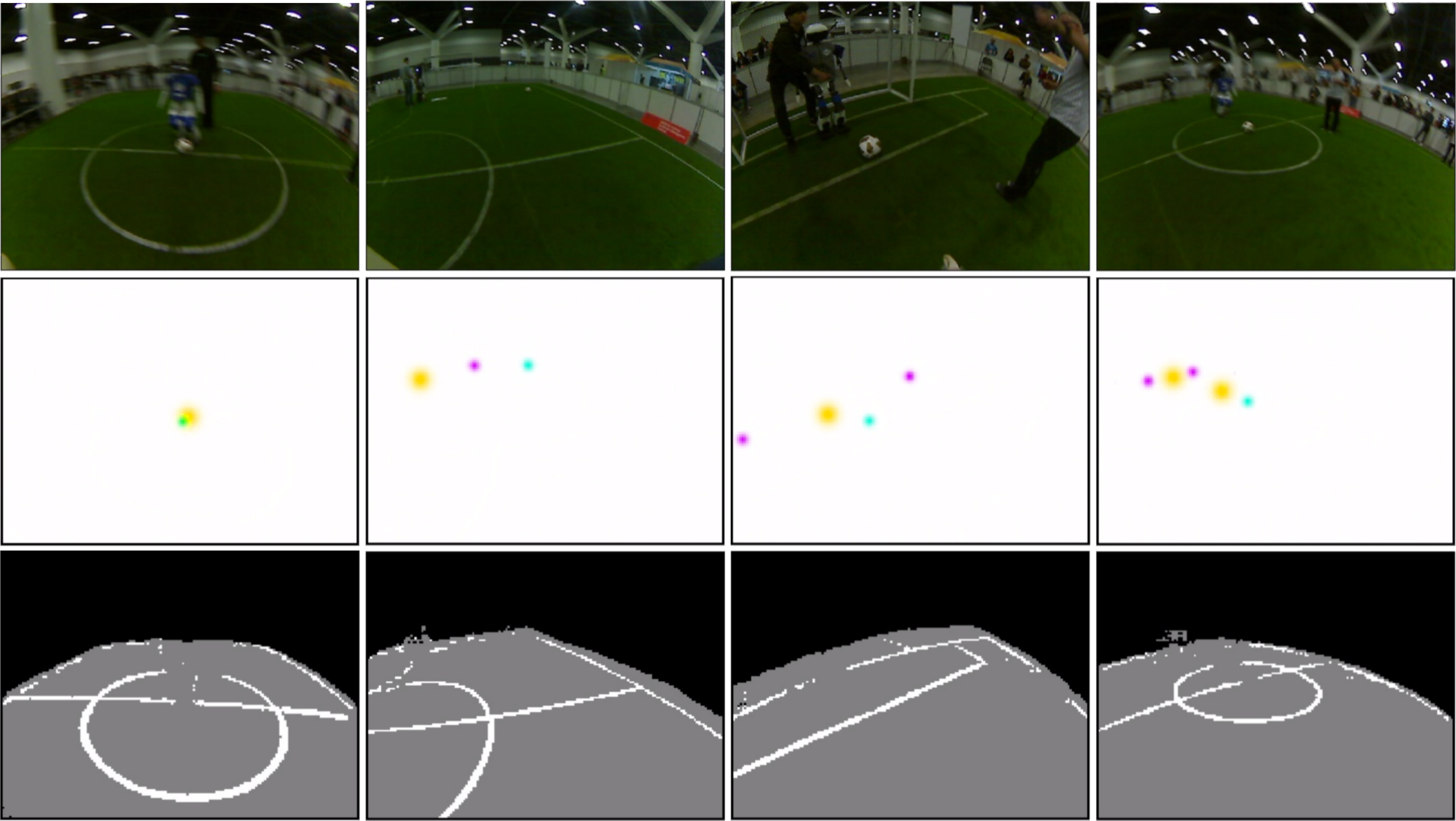}
	\vspace{-3ex}
	\caption{Object detection results. Upper row: captured images by our robots. 
		Middle row: the output of the network with balls (cyan), goal posts (magenta), and robots (yellow). 
		Bottom row: the output of the segmentation branch with lines (white), field (gray), and background (black).}
	\label{vision_output}
	\vspace{-2ex}
\end{figure}

One other difficulty of this year of RoboCup was very thin goalposts which were hard to detect. However, with many training samples, the network finally managed to learn it very robustly.
After sufficient training, goal posts were detected even when they were hard to recognize by a human. This might be explained by inferring their presence from other features of the pitch like field boundary and lines. One detected hard-to-recognize goal post is shown in the last row of Fig.~\ref{vision_output}.

Despite using Adam optimizer, which has an adaptable per-parameter learning rate, 
finding a suitable learning rate is still a challenging prerequisite for training. 
To determine an optimal learning rate, we followed the approach presented by Smith et. al.~\cite{smith2017cyclical}.
Each batch contained only some samples for one of the output heads. We used progressive image resizing that uses small pictures at the beginning of training, 
and step by step increase the dimensions as training progresses, a method inspired by Brock et. al.~\cite{brock2017freezeout} and by Yosinski et. al.~\cite{yosinski2014transferable}. 
In early iterations, the inaccurate randomly initialized model will make fast progress by learning from large batches of small pictures. Within the initial fifty epochs, 
we used downsampled training images, whereas the weights on the encoder part are frozen. Throughout the following fifty epochs, all parts of the models are jointly trained. 
In the last fifty epochs, full-sized pictures are used to learn fine-grained details. 
A lower learning rate is employed for the encoder part, with the intuition that the pre-trained model needs less training time to converge. 
With the described method, the entire training process with around 9k samples takes less than three hours on a single Titan X GPU with \SI{12}{GB} of memory. 
Examples from the test set are pictured in Fig.~\ref{vision_output}. 
To annotate more data as quickly as possible, we designed an annotation tool which automatically annotates the input based on the previously trained model. The user then only had to correct those samples which were wrongly classified. This semi-automatic annotation tool was crucial for us to gather as many samples as possible from the RoboCup 2019 environment.

The output of the network is of lower resolution and has less spatial information than the input image. 
To account for this effect in the detection part, we calculate sub-pixel level coordinates based on the center of mass of a detected contour. 
There was no need to account for lower resolution output in the field and line segmentation.

\begin{figure}[t]
	\vspace{-2ex}
	\centering
	\includegraphics[width=0.8\linewidth]{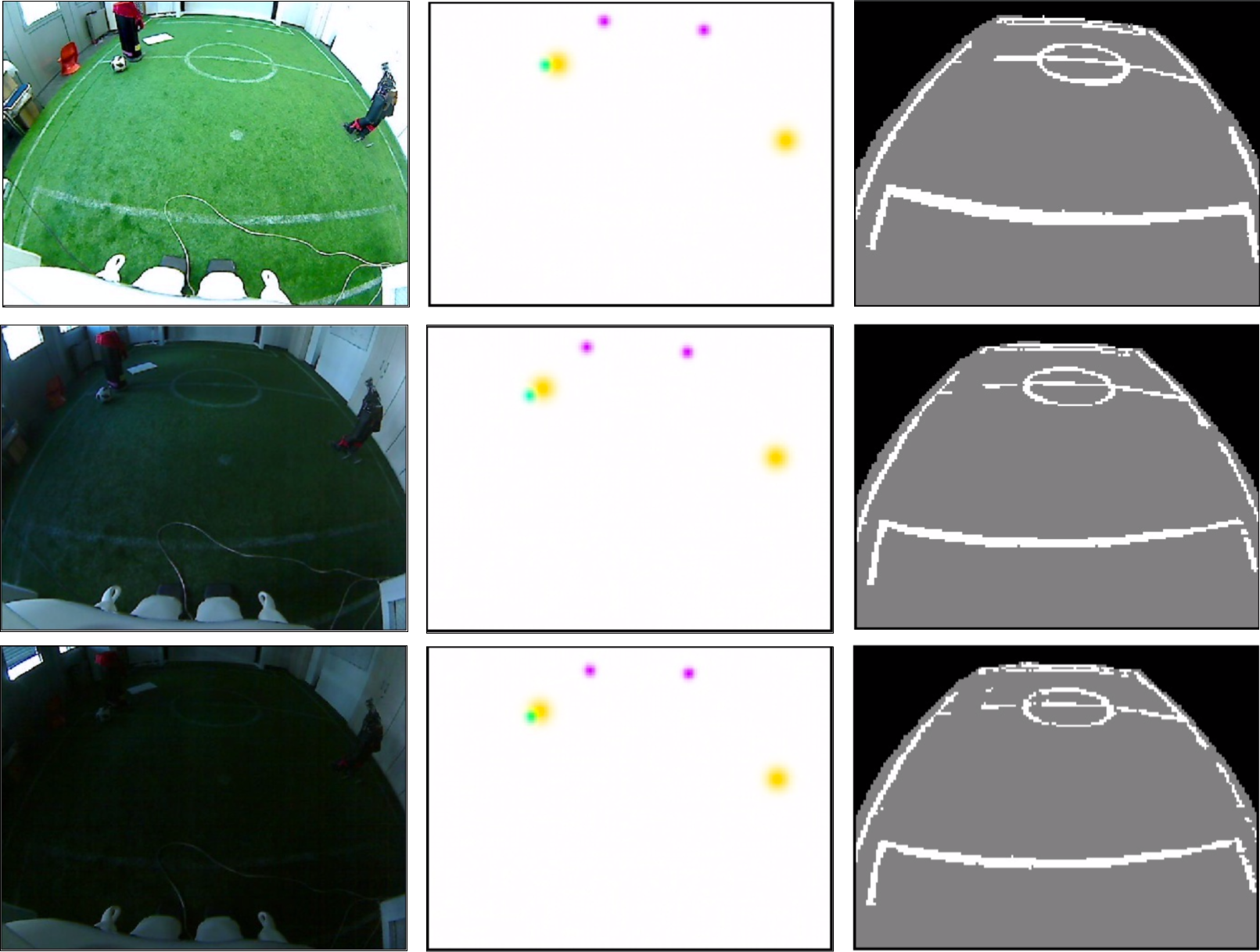}
	\vspace{-1ex}
	\caption{Object detection under various lighting conditions. Left column: captured images by our robots. Middle column: output of the network indicating balls (cyan), goal posts (magenta), and robots (yellow). Right column: output of the segmentation branch showing lines (white), field (gray), and background (black).}
	\label{vision_output_dark}
	\vspace{-3ex}
\end{figure}
After detecting soccer-related objects, we filter them and project each object location into egocentric world coordinates. 
Using NimbRoNet2, we can detect objects which are up to 10 meters away.
The complete perception pipeline, including a forward-pass of the network, takes approximately \SI{36}{ms} on the robot hardware.
Using a unified network helped both detection and segmentation. The network learned to exclude the balls which were outside of the field, hence reducing false detection rate. Outside field object removal was previously done only after post-processing. 
In addition, the robot was able to play soccer in pitch black, and the perception was robust in different lighting conditions, including direct sunlight and without ambient light (Fig.~\ref{vision_output_dark}).
Unfortunately, this year, all AdultSize games were played with artificial light,
thus we could not test our new development for lighting conditions during the competition.

\begin{table}[bt!]
	\caption {Results of the detection branch of our visual perception network.} \label{vis_res_det} 
	\vspace{-3ex}
	\begin{center}
		\begin{tabular}{l | c  c  c  c  c}
			\toprule
			Type & F1 & Accuracy & Recall & Precision & FDR \\
			\midrule
			Ball (NimbRoNet2) & \textbf{0.998} & \textbf{0.996} & 0.996 & \textbf{1.0} & \textbf{0.0} \\
			Ball (NimbRoNet) & 0.997 & 0.994 & \textbf{1.0} & 0.994 & 0.005 \\
			Ball (SweatyNet-1~\cite{schnekenburger2017detection}) & 0.985 & 0.973 & 0.988 & 0.983 & 0.016 \\
			\midrule
			Goal (NimbRoNet2) & \textbf{0.981} & \textbf{0.971} & {0.973} & \textbf{0.988} & \textbf{0.011} \\
			Goal (NimbRoNet) & {0.977} & {0.967} & \textbf{0.988} & {0.966} & {0.033} \\
			Goal (SweatyNet-1~\cite{schnekenburger2017detection}) & 0.963 & 0.946 & 0.966 & 0.960 & 0.039 \\
			\midrule
			Robot (NimbRoNet2) & \textbf{0.979} & \textbf{0.973} & \textbf{0.963} & \textbf{0.995} & \textbf{0.004} \\
			Robot (NimbRoNet) & {0.974} & {0.971} & {0.957} & {0.992} & {0.007} \\
			Robot (SweatyNet-1~\cite{schnekenburger2017detection}) & 0.940 & 0.932 & {0.957} & 0.924 & 0.075 \\
			\midrule
			Total (NimbRoNet2) & \textbf{0.986} & \textbf{0.986} & {0.977} & \textbf{0.994} & \textbf{0.005} \\
			Total (NimbRoNet) & {0.983} & {0.977} & \textbf{0.982} & {0.984} & {0.015} \\
			Total (SweatyNet-1~\cite{schnekenburger2017detection}) & 0.963 & 0.950 & 0.970 & 0.956 & 0.043 \\
			\bottomrule
		\end{tabular}
	\end{center}
	\vspace{-2ex}
\end{table}

Our visual perception pipeline is compared on different soccer-related objects against SweatyNet~\cite{schnekenburger2017detection} and our previous model NimbRoNet~\cite{ficht2018nimbro} (Table.~\ref{vis_res_det}).
We also evaluated our segmentation head (Table.~\ref{vis_res_seg}). 
We have outperformed SweatyNet and NimbRoNet, 
whose results were one of the best-reported in terms of detecting soccer objects. 
This achievement was also accompanied by being approximately two times faster than SweatyNet in training phase. 
The reduced training time can be attributed to the progressive image resizing and transfer learning techniques. 
\begin{table}[tb]
	\caption {Results of the semantic segmentation of our visual perception network.} \label{vis_res_seg} 
	\vspace{-3ex}
	\begin{center}
		\begin{tabular}{ l | c  c}
			\toprule
			Type & Accuracy & IOU \\
			\midrule
			Field & {0.986} & {0.975}  \\
			Lines & {0.881} & {0.784}  \\
			Background & {0.993} & {0.981}  \\
			\midrule
			Total & {0.953} & {0.913}  \\
			\bottomrule
		\end{tabular}
	\end{center}
	\vspace{-3ex}
\end{table}
\section{Robust Omnidirectional Gait with In-walk Kick}
Team NimbRo has developed a motion and a gait control framework capable of absorbing pushes from any direction at any time during the gait cycle. 
This year, 
for the first time,
our NimbRo-OP2(X) adult-sized platforms incorporated step-based push recovery capabilities and in-walk kicks.
\subsubsection{Compliant Actuation.}
Motions performed by the robot are sensitive to the tracking capabilities of the control system. 
We developed a feed-forward control scheme which modifies the joint trajectories based on the commanded position and inverse dynamics~\cite{Schwarz2013a}. 
The model incorporates factors such as battery voltage, joint frictions, and body inertias.
\subsubsection{Open-loop Walking.}
The walking gait is based on an open-loop central pattern generator calculated from a gait phase angle proportional to the desired gait frequency.
To formulate this open-loop gait, we use three different spaces: joint space, Cartesian space and abstract space \cite{Behnke2006}.
The open-loop gait is further extended by the integration of an explicit double support phase, modification of the leg extension profiles, and velocity and acceleration-based leaning strategies~\cite{Allgeuer2016a}.
These extensions resulted in passive damping of oscillations and smooth transition between swing and support phases.
\subsubsection{Feedback Mechanisms.}
Several basic feedback mechanisms, namely arm angle, hip angle, continuous foot angle, support foot angle, CoM shifting, and virtual slope, have been built around the open-loop gait core to stabilize the walking~\cite{Allgeuer2016a}. 
These PID-like feedback mechanisms derive from the state estimation and add corrective action components to the central pattern generated waveforms.
\subsubsection{Capture Steps Gait.}
We use the Capture Step Framework~\cite{missura2016analytic,missura2014balanced} to make our robots recover balance.
The Capture Step Framework is a composition of central pattern-generated open-loop
step motions and a linear inverted pendulum model-based balance controller. In each
iteration of the motion control loop, timing and location of the next footstep
are computed using the linearized equations of the inverted pendulum model such that
the Center of Mass (CoM) would return to a stable limit cycle while also following a commanded
walking velocity. Our robots showed stable walking throughout the competition, including
balance-restoring capture steps after collisions in games and excelled in the technical
challenge.

\subsection{In-walk Kick}
\label{sec:in_walk_kick}
\begin{figure}[b]
	\centering
	\includegraphics[height=0.275\linewidth]{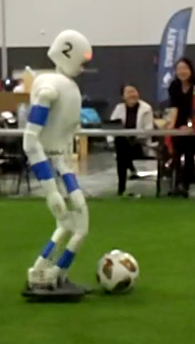}
	\includegraphics[height=0.275\linewidth]{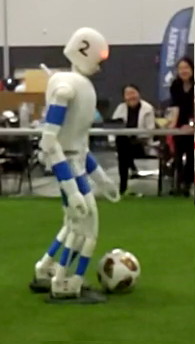}
	\includegraphics[height=0.275\linewidth]{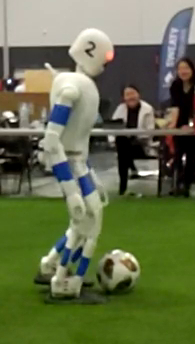}
	\includegraphics[height=0.275\linewidth]{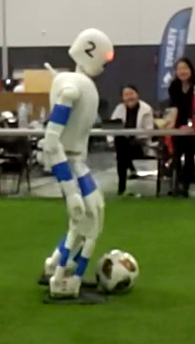}
	\includegraphics[height=0.275\linewidth]{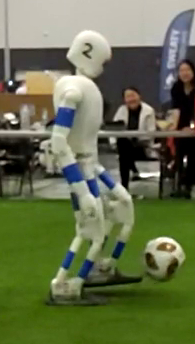}
	\includegraphics[height=0.275\linewidth]{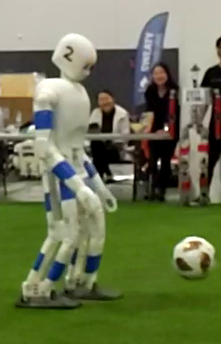}
	\caption{In-walk kick performed during a soccer match.}
	\label{fig:ok_execution}
\end{figure}
Our in-walk kick approach integrates the kick directly into the gait to avoid unnecessary stops (Fig.~\ref{fig:ok_execution}),
which were required in our previous kicking motions.
Thus, a significantly boosting of the overall pace of the game on the larger field has been achieved.

\begin{figure}[t]
	\centering
	\includegraphics[width=0.8\textwidth]{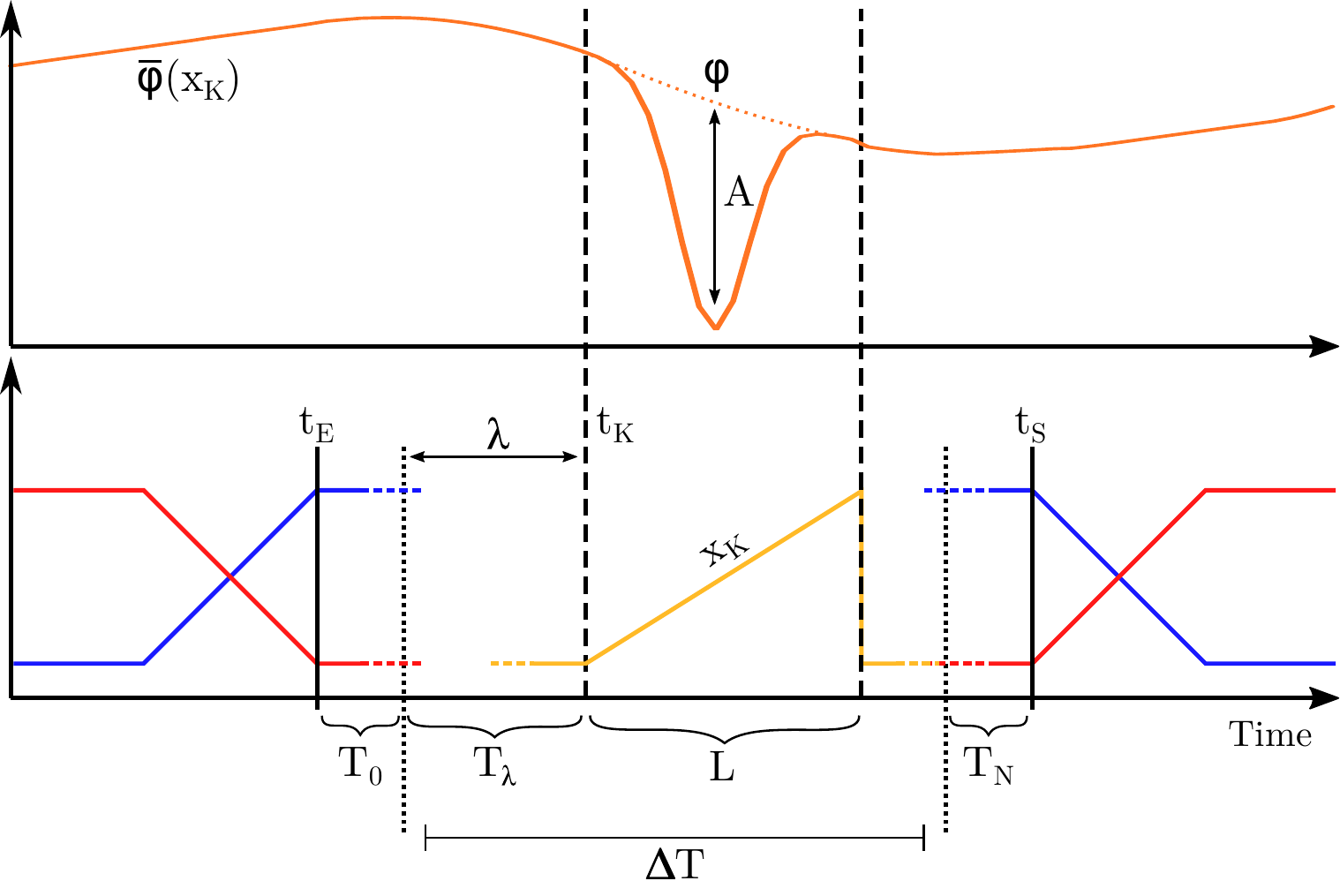} 
	\caption{A schematic visualization of the kick phase $x_K$ and the sagittal leg angle $\bar{\varphi}$. 
		The upper part shows the augmented leg angle $\bar{\varphi}$ in sagittal 
		direction during a kick, where the vertical dashed lines symbolize the start and the end 
		of the kicking motion. The lower part displays the timing parameters used to calculate $x_K$, where the red line resembles the support coefficient of the kicking leg and the blue line of the supporting leg respectively. The timing parameter $\lambda$ can be used to move the motion of length $L$ inside the legal execution window of size $\Delta T$.}
	\label{fig:ok_schematic}
\end{figure}

A general schematic of the approach can be seen in Fig.~\ref{fig:ok_schematic}. 
In our approach, an allowed time window $\Delta T$ is defined for the kick.
Generally, a kick can be performed between the end $t_E$ of the previous 
support transition phase and the start $t_S$ of the next transition phase respectively. 
Nevertheless, for $T_0,T_N>0$, it is advantageous to prohibit kicks in boundary 
intervals $[t_s, t_s+T_0]$ and $[t_E-T_N, t_E]$ to prevent unwanted foot contact with the 
ground during the kick execution.
In this manner, we define 
\begin{equation}
 \Delta T = t_E - t_S - T_0 - T_N
\end{equation}
as the length of the interval where we can safely perform a kick. In addition, given a 
motion duration $L<\Delta T$, it is possible to perform the kick inside an 
arbitrary location of the allowed interval. This enables us to define a timing 
parameter $0\leq\lambda\leq1$ resulting in a delay interval 
\begin{equation}
 T_\lambda = \lambda\left(\Delta T -L\right)\,,
\end{equation}
controlling the starting time of motion execution inside the allowed time window. 
This is particularly important for the soccer behaviors, since it allows for exact control 
of the location of the foot at the start and apex of the kick. Altogether, this results 
in the actual starting time of the kick 
\begin{equation}
 t_K = t_S+T_0+T_\lambda\,.
\end{equation}

Consequently, a kick phase $x_K$ is defined, which linearly 
interpolates from $-1$ to $1$ between the nominal start and end of the kicking 
motion: 
\begin{equation}
 x_K\left(t\right) = 
   2\frac{t-\left(t_S+T_0+T_\lambda\right)}{L}-1.
\end{equation}
In the end, the kick phase is used to compute the augmented sagittal leg angle:
\begin{equation}
 \bar{\varphi}\left(t\right) =
 \begin{cases}
   \varphi - A\exp{\left(-\frac{1}{2}\left(\frac{x_K\left(t\right)}{\sigma}\right)^2\right)}, & 
   \text{if } t \in [t_K,t_K + L] \\
   \varphi, & \text{otherwise}
  \end{cases}
\end{equation}
by subtracting a Gaussian curve of the sagittal leg motion, where $A$ defines the amplitude and $\sigma$ 
controls the width of the Gaussian, achieving a smooth but distinct forward 
motion of the leg. The Gaussian term is set to zero 
beyond the interval boundaries $[t_K,t_K + L]$. Thus, $\sigma$ has to be small 
enough such that the activation of the Gaussian can be neglected at the borders, 
ensuring that the transition to the kick is smooth.

\section{Soccer Behaviors}
We refer to soccer behaviors to the decision process required to play football.
These decisions include, for example, to search for the ball if this is not detected,
to go for the ball if we are far from it,
or to activate the kick if all the conditions for kicking are granted.
The decision process is modeled as a hierarchical Finite State Machine (FSM) with two main layers~\cite{rodriguez2017advanced}.
The state of the upper layer is established by the state of the game defined in part by the game controller and the role of the player (goalie, striker or defender).
The states of the lower FSM represent individual skills of the players such as: move, stop, kick, dribble, dive, among others.
Collision avoidance, i.e., avoidance of other robots---either from the opponent team or our team---is part of the lower state machine.

\subsubsection{Team Play Strategies.}
In general, the function of the team play is to safely assign the game roles to each of the players.
In this manner, for example, having two strikers simultaneously is not desired in order to avoid collisions between robots of the same team when they are going for the ball.
The task assignment is implemented as a server/client architecture where the striker is the server and it is the only one allowed to accept task renegotiation requests.
The other players, i.e., defenders and goalies, are allowed to make requests if they find themselves in a better position than the striker, e.g., being closer to the ball.
During drop-in games, no task renegotiation was allowed, 
mainly due to the lack of game roles of other teams.
Thus, our players were assigned to fixed roles from the beginning of the match. 
However,
team capabilities were still exhibited during drop-in games, 
e.g., by our goalkeepers clearing out balls and returning to their corresponding goal. 
For a deeper discussion about the team play strategies, please refer to~\cite{rodriguez2017advanced}.

\section{Technical Challenges}
\seclabel{technical_challenges}
Technical challenges is a separate competition, where robots have to perform isolated independent tasks during a limited time period.
Since the time period for executing all tasks is limited to only 25 minutes, 
robustness and reliability have the highest importance when designing a solution for each challenge. 
At RoboCup~2019 there were four technical challenges: push recovery, high jump, high kick and goal kick from moving ball. 

\subsection{Push Recovery}
\begin{figure}
	\centering
	\includegraphics[width=0.24\linewidth]{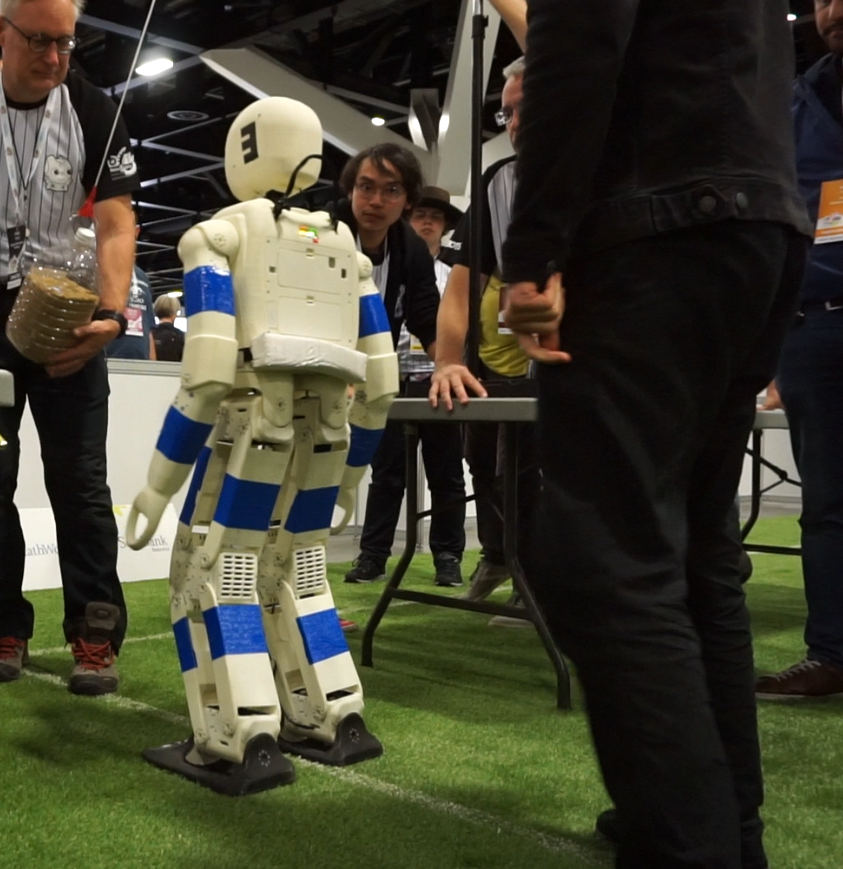}
	\includegraphics[width=0.24\linewidth]{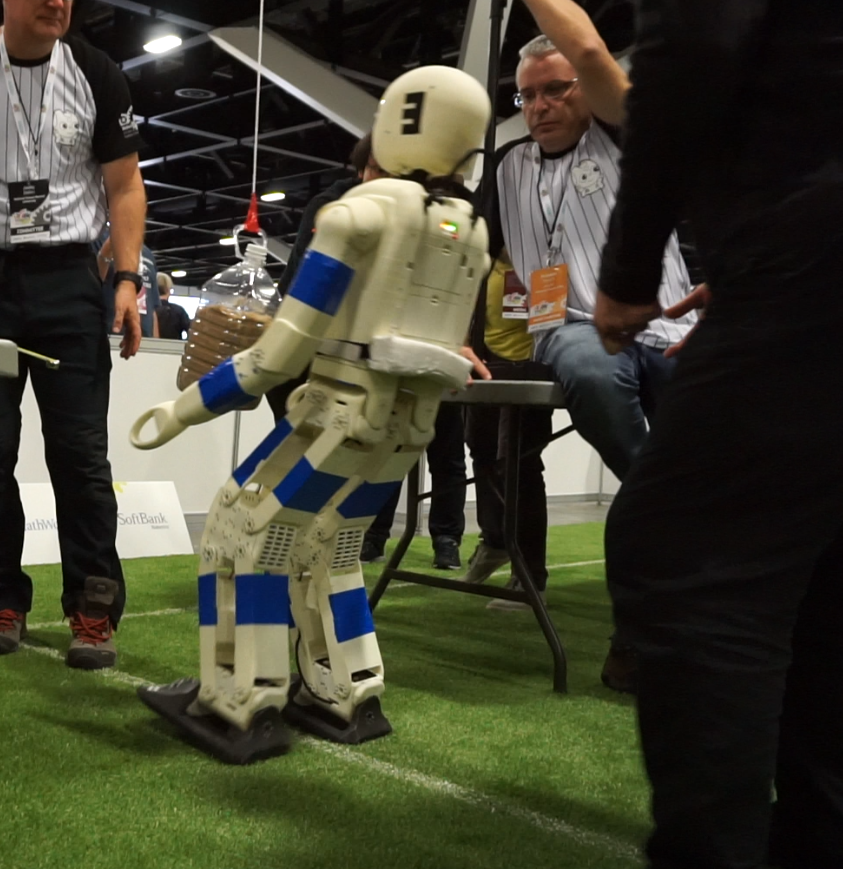}
	\includegraphics[width=0.24\linewidth]{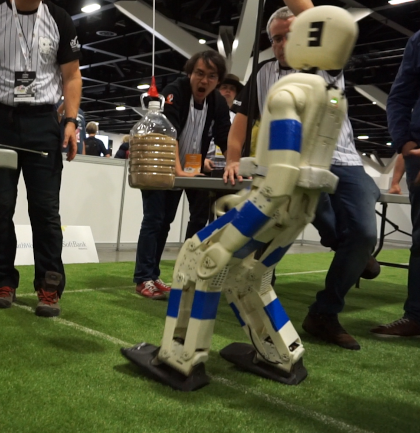}
	\includegraphics[width=0.24\linewidth]{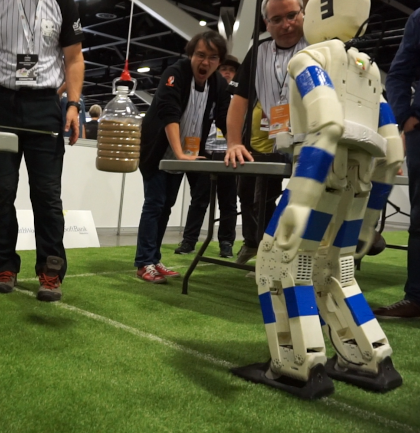}
	\caption{Technical Challenge: Push Recovery. Several capture steps allow the robot to regain balance after a very strong push of a 5\,kg pendulum.}
	\label{fig:push_recovery}
	\vspace*{-4ex}
\end{figure}

\seclabel{push_recovery}
In this \textit{Push Recovery} challenge, a robot has to withstand three pushes in a row while walking on spot. 
The pushes are performed by releasing a previously retracted pendulum which then hits the robot at the height of the CoM. 
The pushes are performed randomly from the front and from the back. 
The weight of the pendulum is 5~kg and the robots are ranked by the distance of pendulum retraction for the series of three successful attempts. 
The Capture Step Framework allowed our robot to withstand very strong pushes, making a series of capture steps to regain balance~(Fig.~\ref{fig:push_recovery}), and to finish first in this challenge.

\subsection{High Jump}
\seclabel{high_jump}
The goal of the \textit{High Jump} is to remain airborne during a vertical jump as long as possible and upon landing remain in a stable sitting or standing posture. 
For this challenge, motions were pre-designed using key-frames and a simple geometry-based mass distribution principle~\cite{ficht2018online}. 
By lowering and rapidly lifting the CoM, the accumulated linear momentum at full CoM height propels the robot into the air. 
After the leap is performed, there is a possibility to decrease the CoM height, which results in folding the legs. 
This would increase the time in the air by postponing the contact with the ground plane even with a weaker jump upwards. 
However, we observed that this rapid leg movement caused the robot to often lose balance upon landing.
Due to the bent knees, the force upon impact was also damaging to the gears in the knee actuators. In our experience, we have found that landing 
on extended legs increased the durability of the actuators and made the landing more reliable. This was largely due to the integrated 
tension springs in the legs. They provide passive compliance during landing and also contributed greatly to the strength of the jump.
Our robot remained airborne for \unit{0.262}{s} and came in second with a difference of \unit[13]{ms} to the first. 

\subsection{High Kick}
\seclabel{high_kick}
In the \textit{High Kick} challenge, the robot has to score a goal over an obstacle which is positioned on the goal line. 
The ball is initially positioned at the penalty mark. The goal is only valid if the ball surpassed the obstacle without touching it. The height of the obstacle can be adjusted and the teams are ranked by the height of the successfully over-kicked obstacle. Since it is allowed to touch the ball multiple times, we first move the ball close to the goal line by executing a pre-designed kicking motion. 
Having the ball close to the obstacle, we execute a pre-designed high kick motion. During this motion the tip of the foot makes first contact with the ball as close to the ground plane as possible. 
From that point, the foot moves forward and upwards.
In order to improve the efficiency of this motion, we use a modified foot with a "scoop" shape.
It ensures a prolonged contact with the ball during the high kick motion and---hence---transfer of more energy to the ball, which allows to kick over higher obstacles. Our team came in second in this challenge, successfully  kicking over an obstacle of \unit[26]{cm} height.
\begin{figure}[b!]
	\centering
	\includegraphics[width=0.24\linewidth]{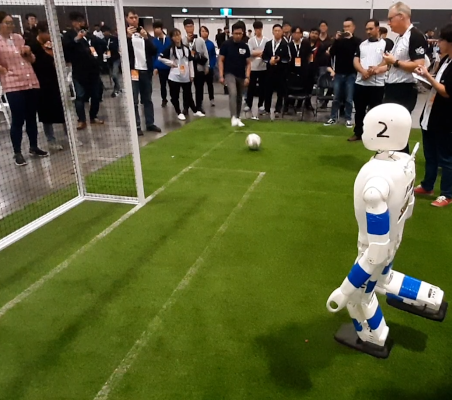}
	\includegraphics[width=0.24\linewidth]{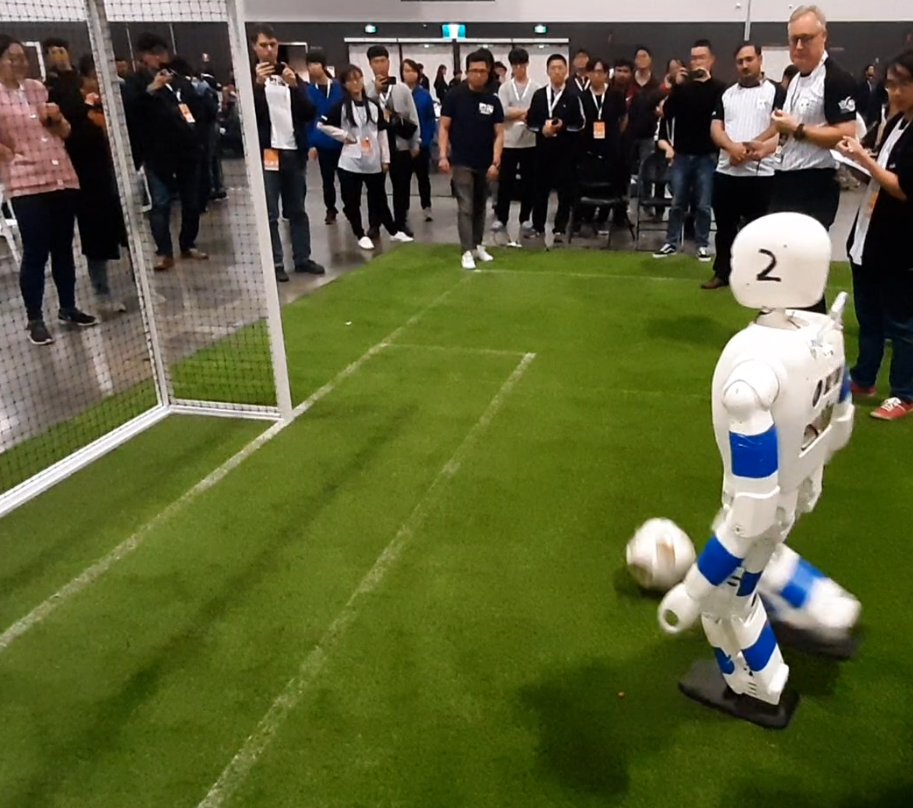}
	\includegraphics[width=0.24\linewidth]{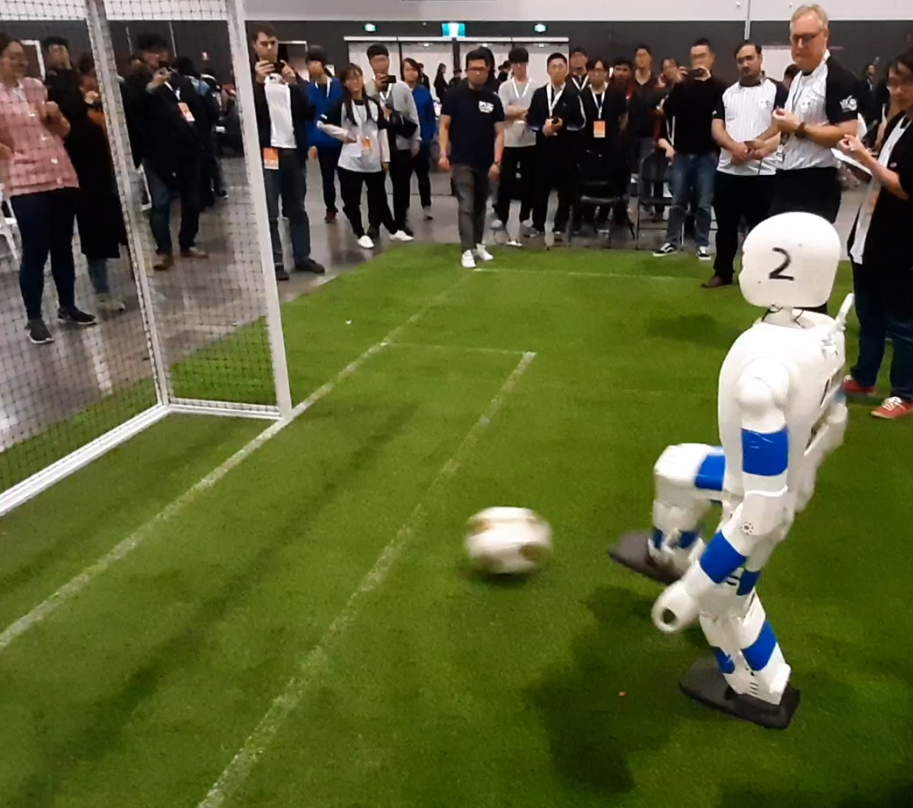}
	\includegraphics[width=0.24\linewidth]{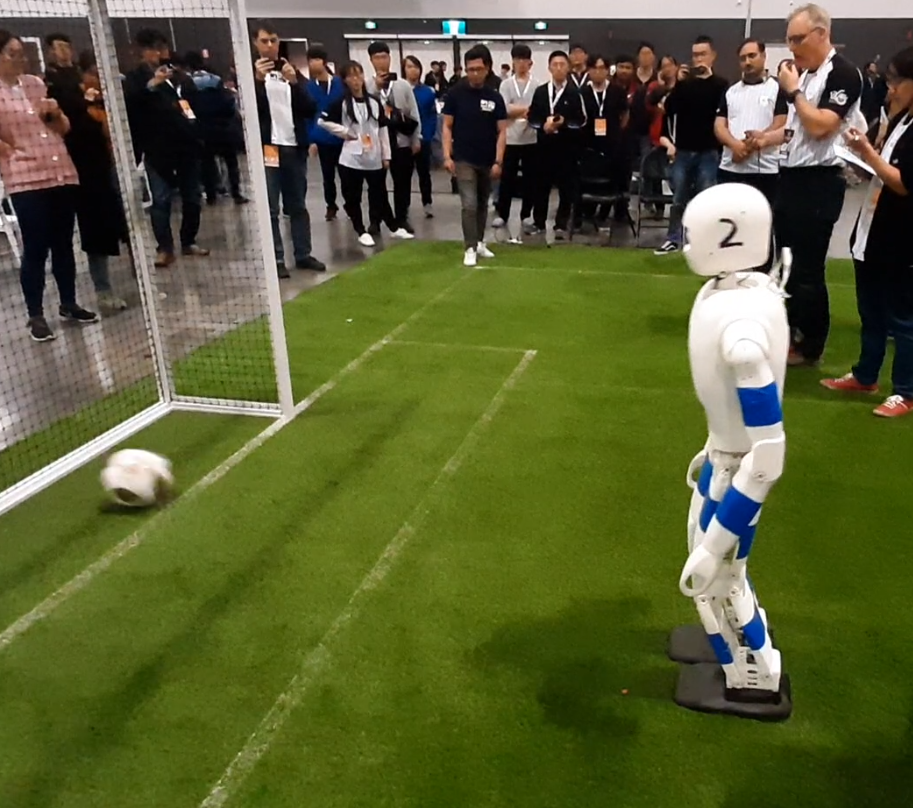}
	\caption{Technical Challenge: Goal Kick from Moving Ball.}
	\label{fig:moving_ball}
\end{figure}

\subsection{Goal Kick from Moving Ball}
\seclabel{moving_ball}
The goal of this challenge is to score a goal by kicking a moving ball.
The robot is placed at the penalty mark. The ball is positioned at the corner of the field and is passed to the robot either by a human or by another robot. 
The teams are ranked by the number of successful goals out of three consecutive attempts. 
In order to know when the robot has to kick, we predicted the time-of-travel of the ball to get in front of the robot foot by estimating its velocity and acceleration from a series of consecutive ball detections, 
separated in time by a time interval $\epsilon = 0.1$~s. 
Our robot performing this challenge is shown in Fig.~\ref{fig:moving_ball}. Our team took the first place in this challenge, successfully scoring the goal from the moving ball three out of three times.

\section{Game Performance}
\seclabel{op2}
During the AdultSize 2 vs. 2 Soccer competition of RoboCup~2019,
our robots scored 48 goals while receiving none. The robots have shown outstanding performance during the whole tournament including winning the final game 8:0.
While 2 vs. 2 competition games have shown individual and team capabilities, drop-in games demonstrated individual skills of each single robot.
In the Drop-in tournament, our robots scored 31:7 goals in 6 drop-in games---resulting in winning 57 points with a margin of 33 points to the second best team.
Compared to the soccer tournament, we received goals during drop-in games mainly due to the lack of a second field player (our partner teams normally placed goalkeepers), 
and due to the lack of diving motions when our robots were goalkeepers.
The capabilities of our robots were once more demonstrated by winning the AdultSize Technical Challenges.
Consequently, NimbRo received the 2019 Best Humanoid Award of the Humanoid League.
\section{Conclusions}
In this paper, we presented the approaches that lead us to win all possible competitions in the AdultSize class for the RoboCup 2019 Humanoids League in Sydney: soccer tournament, drop-in games and technical challenges.
Special emphasis was put on the deep learning based computer vision system that lead our robots to be robust against different lighting conditions and to detect reliably balls up to \unit[10]{m}.
Part of our success in the games was explained by the novel in-walk kick,
making our games very dynamic and hard to counteract by the opponents.
We also presented a step-based push recovery approach that was demonstrated during the competition with impressive performance.
Finally, the decision making process and team play strategies were presented, 
which are responsible for integrating and making use of all individual components.

\subsection*{Acknowledgements}
\footnotesize
This work was partially funded by grant BE 2556/13 of German Research Foundation.


\renewcommand{\bibsection}{\section*{References}}

\begingroup
\small 
\bibliography{winners_2019.bib}

\begin{thebibliography}{21}
\providecommand{\natexlab}[1]{#1}
\providecommand{\url}[1]{\texttt{#1}}
\providecommand{\urlprefix}{}

\bibitem[{Allgeuer and Behnke(2016)}]{Allgeuer2016a}
Allgeuer, P., Behnke, S.: Omnidirectional bipedal walking with direct fused
  angle feedback mechanisms.
\newblock In: 16th IEEE-RAS Int. Conf. on Humanoid Robots (Humanoids) (2016)

\bibitem[{Badrinarayanan et~al.(2015)Badrinarayanan, Kendall, and
  Cipolla}]{badrinarayanan2015segnet}
Badrinarayanan, V., Kendall, A., Cipolla, R.: Seg{N}et: A deep convolutional
  encoder-decoder architecture for image segmentation.
\newblock IEEE Transactions on Pattern Analysis and Machine Intelligence
  39(12), 2481--2495 (2015)

\bibitem[{Behnke(2006)}]{Behnke2006}
Behnke, S.: Online trajectory generation for omnidirectional biped walking.
\newblock In: Proceedings of 2006 IEEE Int. Conf. on Robotics and Automation
  (ICRA) (2006)

\bibitem[{Brock et~al.(2017)Brock, Lim, Ritchie, and
  Weston}]{brock2017freezeout}
Brock, A., Lim, T., Ritchie, J.M., Weston, N.: Freezeout: Accelerate training
  by progressively freezing layers.
\newblock arXiv:1706.04983  (2017)

\bibitem[{Farazi et~al.(2015)Farazi, Allgeuer, and Behnke}]{farazi2015}
Farazi, H., Allgeuer, P., Behnke, S.: A monocular vision system for playing
  soccer in low color information environments.
\newblock In: 10th Workshop on Humanoid Soccer Robots, IEEE-RAS Int. Conf. on
  Humanoid Robots (2015)

\bibitem[{Farazi and Behnke(2016)}]{farazi2016real}
Farazi, H., Behnke, S.: Real-time visual tracking and identification for a team
  of homogeneous humanoid robots.
\newblock In: RoboCup 2016: Robot World Cup XX. pp. 230--242. Springer (2016)

\bibitem[{Farazi and Behnke(2017)}]{Farazi2017b}
Farazi, H., Behnke, S.: Online visual robot tracking and identification using
  deep {LSTM} networks.
\newblock In: IEEE/RSJ Int. Conf. on Intelligent Robots and Systems (IROS)
  (2017)

\bibitem[{Farazi et~al.(2019)Farazi, Ficht, Allgeuer, Pavlichenko, Rodriguez,
  Brandenburger, Hosseini, and Behnke}]{Farazi2019NimbRo}
Farazi, H., Ficht, G., Allgeuer, P., Pavlichenko, D., Rodriguez, D.,
  Brandenburger, A., Hosseini, M., Behnke, S.: {NimbRo Robots Winning RoboCup
  2018 Humanoid AdultSize Soccer Competitions}.
\newblock RoboCup 2018: Robot World Cup XXII, Lecture Notes in Computer Science
   (2019)

\bibitem[{Ficht et~al.(2017)Ficht, Allgeuer, Farazi, and
  Behnke}]{ficht2017nop2}
Ficht, G., Allgeuer, P., Farazi, H., Behnke, S.: {N}imb{R}o-{O}{P}2: Grown-up
  3{D} printed open humanoid platform for research.
\newblock In: 17th IEEE-RAS Int. Conf. on Humanoid Robots (Humanoids) (2017)

\bibitem[{Ficht and Behnke(2018)}]{ficht2018online}
Ficht, G., Behnke, S.: Online {B}alanced {M}otion {G}eneration for {H}umanoid
  {R}obots.
\newblock In: IEEE-RAS 18th Int. Conf. on Humanoid Robots (Humanoids) (2018)

\bibitem[{Ficht et~al.(2018{\natexlab{a}})Ficht, Farazi, Brandenburger,
  Rodriguez, Pavlichenko, Allgeuer, Hosseini, and Behnke}]{ficht2018nimbro}
Ficht, G., Farazi, H., Brandenburger, A., Rodriguez, D., Pavlichenko, D.,
  Allgeuer, P., Hosseini, M., Behnke, S.: Nimb{R}o-{O}{P}2{X}: Adult-sized
  open-source 3{D} printed humanoid robot.
\newblock In: 18th IEEE-RAS Int. Conf. on Humanoid Robots (2018{\natexlab{a}})

\bibitem[{Ficht et~al.(2018{\natexlab{b}})Ficht, Pavlichenko, Allgeuer, Farazi,
  Rodriguez, Brandenburger, Kuersch, and Behnke}]{ficht2018Grown}
Ficht, G., Pavlichenko, D., Allgeuer, P., Farazi, H., Rodriguez, D.,
  Brandenburger, A., Kuersch, J., Behnke, S.: {Grown-up {N}imb{R}o Robots
  Winning {R}obo{C}up 2017 Humanoid {A}dult{S}ize Soccer Competitions}.
\newblock In: RoboCup 2017: Robot World Cup XXI. pp. 448--460. Springer
  (2018{\natexlab{b}})

\bibitem[{Missura and Behnke(2014)}]{missura2014balanced}
Missura, M., Behnke, S.: Walking with capture steps.
\newblock In: IEEE-RAS Int. Conf. on Humanoid Robots. pp. 526--526 (2014)

\bibitem[{Missura(2016)}]{missura2016analytic}
Missura, M.: Analytic and learned footstep control for robust bipedal walking.
\newblock Ph.D. thesis, Universit{\"a}ts-und Landesbibliothek Bonn (2016)

\bibitem[{{Niloofar Azizi, Hafez Farazi} and Behnke(2018)}]{AziziFarazi2018}
{Niloofar Azizi, Hafez Farazi}, Behnke, S.: Location dependency in video
  prediction.
\newblock In: Int. Conf. on Artificial Neural Networks (ICANN) (2018)

\bibitem[{Rodriguez et~al.(2017)Rodriguez, Farazi, Allgeuer, Pavlichenko,
  Ficht, Brandenburger, K{\"u}rsch, and Behnke}]{rodriguez2017advanced}
Rodriguez, D., Farazi, H., Allgeuer, P., Pavlichenko, D., Ficht, G.,
  Brandenburger, A., K{\"u}rsch, J., Behnke, S.: Advanced soccer skills and
  team play of {RoboCup 2017 TeenSize} winner {NimbRo}.
\newblock In: RoboCup 2017: Robot World Cup XXI. pp. 435--447. Springer (2017)

\bibitem[{Ronneberger et~al.(2015)Ronneberger, Fischer, and
  Brox}]{ronneberger2015u}
Ronneberger, O., Fischer, P., Brox, T.: U-{N}et: Convolutional networks for
  biomedical image segmentation.
\newblock In: Int. Conf. on Medical Image Computing and Computer-Assisted
  Intervention (MICCAI). Springer (2015)

\bibitem[{Schnekenburger et~al.(2017)Schnekenburger, Scharffenberg, W{\"u}lker,
  Hochberg, and Dorer}]{schnekenburger2017detection}
Schnekenburger, F., Scharffenberg, M., W{\"u}lker, M., Hochberg, U., Dorer, K.:
  Detection and localization of features on a soccer field with feedforward
  fully convolutional neural networks ({FCNN}) for the {A}dult-size humanoid
  robot {S}weaty.
\newblock In: Proceedings of the 12th Workshop on Humanoid Soccer Robots,
  IEEE-RAS Int. Conf. on Humanoid Robots (Humanoids). (2017)

\bibitem[{Schwarz and Behnke(2013)}]{Schwarz2013a}
Schwarz, M., Behnke, S.: Compliant robot behavior using servo actuator models
  identified by iterative learning control.
\newblock In: {RoboCup} 2013: Robot World Cup {XVII} (2013)

\bibitem[{Smith(2017)}]{smith2017cyclical}
Smith, L.N.: Cyclical learning rates for training neural networks.
\newblock In: Applications of Computer Vision (WACV), IEEE Winter Conf. on. pp.
  464--472 (2017)

\bibitem[{Yosinski et~al.(2014)Yosinski, Clune, Bengio, and
  Lipson}]{yosinski2014transferable}
Yosinski, J., Clune, J., Bengio, Y., Lipson, H.: How transferable are features
  in deep neural networks?
\newblock In: Advances in Neural Information Processing Systems (NIPS). pp.
  3320--3328 (2014)

\end{thebibliography}
\endgroup

\end{document}